%% file: main.tex
\documentclass[runningheads]{llncs}

\usepackage{eccv}

\usepackage{epsfig}
\usepackage{soul}
\usepackage{url}
\usepackage[hidelinks]{hyperref}
\usepackage[utf8]{inputenc}
\usepackage{amsmath}
\usepackage{booktabs}
\usepackage{algorithm}
\usepackage{algorithmic}
\usepackage{multirow}
\usepackage{bm}
\usepackage[table]{xcolor}
\usepackage{pifont}
\usepackage{makecell}

\hypersetup{
    colorlinks,
    citecolor=cyan,
    linkcolor=red, 
    urlcolor=cyan 
}

\usepackage{eccvabbrv}

\usepackage{graphicx}
\usepackage{booktabs}

\usepackage[accsupp]{axessibility}  

\usepackage{hyperref}

\usepackage{orcidlink}

\begin{document}

\title{IMMoE: Incomplete Multi-View Anomaly Detection via Mixture of View Experts Fusion} 

\titlerunning{IMMoE}

\author{Lei Hu \orcidlink{0000-0002-6884-2912}}

\authorrunning{Lei Hu}

\institute{South China University of Technology, Guangzhou, China \\
\email{hulei777@foxmail.com}}

\maketitle

\begin{abstract}

Existing Multi-view Anomaly Detection (MAD) methods assume that all views are completely available and model each view separately. However, in real industrial scenarios, information in the view may be missing due to faults such as occlusion, which leads to the performance degradation of existing methods due to the lack of a multi-view consistency prior. To address this, we explored a more challenging task: Incomplete Multi-View Anomaly Detection (IMVAD), in which some areas of each view were masked. We proposed a pipeline for automatically generating the IMVAD dataset and generated the \textbf{RIMAD} dataset based on the Real-IAD dataset through this pipeline. In addition, in order to effectively utilize the information of multiple views in the absence of view information, we propose \textbf{IMMoE}, which consists of two key modules: (1) Multi-View Expert Fusion (MVEF) effectively fuses multi-view information through a multi-view expert network and guides the reconstruction of a single view; (2) Local Anomaly Enhancement Encoder (LAEE) effectively prevents the model from overfitting the mask region by applying dropout to local features. Our method achieves state-of-the-art performance on both the RIMAD and Real-IAD datasets, especially on RIMAD, we have increased the pixel-level and image-level metrics by 11.8\% and 2.8\%, respectively. Our source code is available at \url{https://github.com/HULEI7/IMMoE}.
\keywords{Anomaly Detection \and Incomplete Multi-View \and Mixture of Experts}
\end{abstract}

\begin{figure}[t]
\centerline{\epsfig{figure=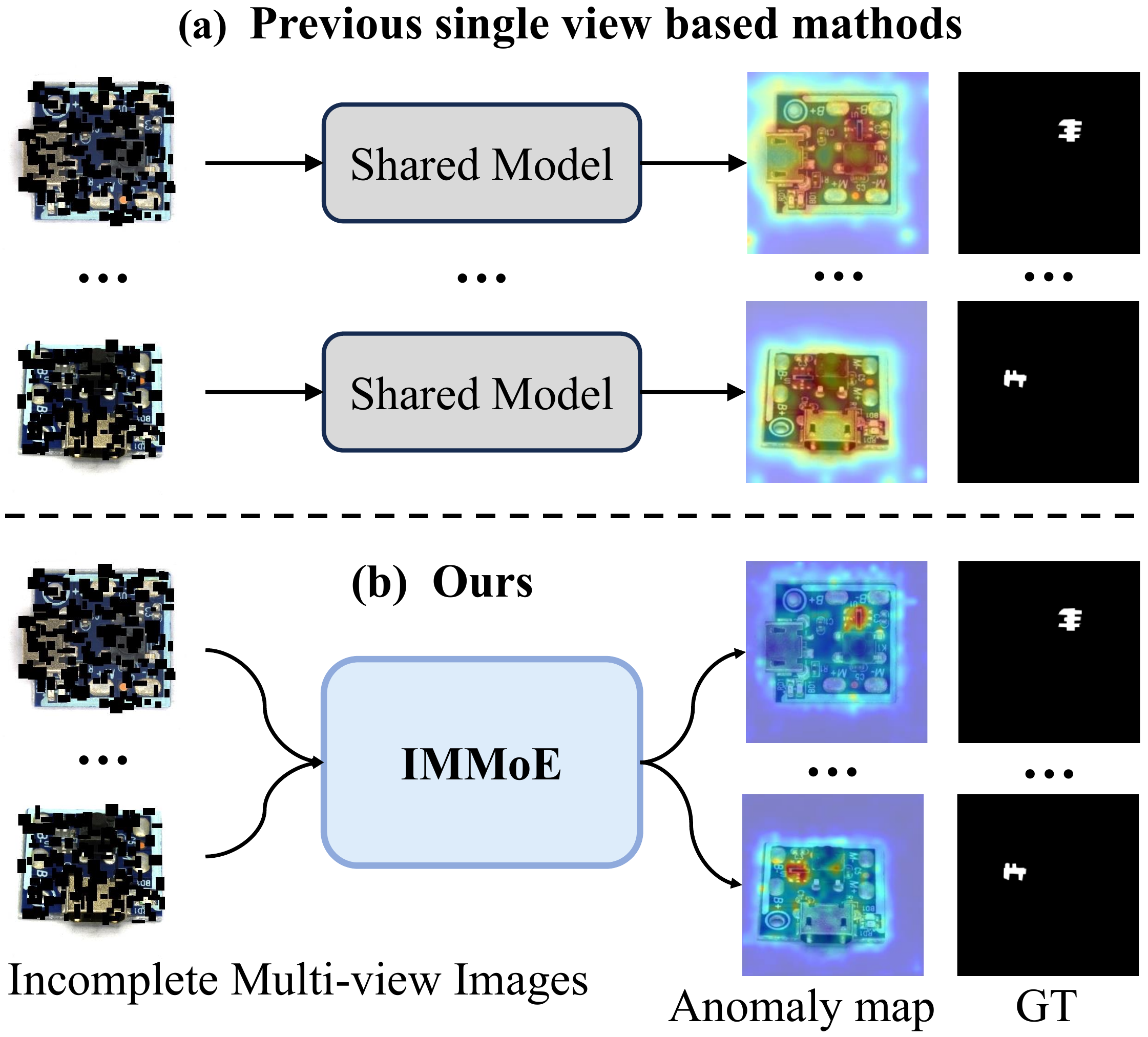,width=8cm}}
\caption{Comparisons of our method with the previous single-view-based method on IMVAD. (a) Due to the lack of multi-view interaction, previous methods will experience significant performance degradation when partial information is missing. (b) Our method effectively integrates the useful information among various views through a multi-view expert network, thereby achieving effective information complementarity. See Figure \ref{fig:4} for more comparisons.}
\label{fig:1}
\end{figure}

\input{section/Introduction}
\input{section/Related_Work}

\input{section/IMVAD}
\input{section/Methods}
\input{section/Experiments}

\section{Conclusion}
This paper introduces IMVAD (Incomplete Multi-View Anomaly Detection), a challenging task reflecting real-world view incompleteness in industrial anomaly detection. To address this, we construct the RIMAD benchmark and propose the IMMoE framework. Our method integrates a Local Anomaly Enhancement Encoder (LAEE) to simulate anomaly patterns, a Multi-view Expert Fusion (MVEF) module to capture cross-view correlations for information reconstruction, and an Area Adaptive Loss (AAL) to prioritize challenging regions. This synergistic design maintains inter-view consistency and ensures robust detection even under severe view-incompleteness. Experimental results and heatmaps demonstrate that IMMoE achieves state-of-the-art performance on both RIMAD and Real-IAD datasets. Especially on RIMAD, we have increased the pixel-level and image-level metrics by 11.8\% and 2.8\%, respectively. This work underscores the necessity of multi-view synergy and provides a robust solution for comprehensive industrial inspection under incomplete information.

\clearpage

%
%
\bibliographystyle{splncs04}
\bibliography{main}

\input{section/sup}

\end{document}

%% file: section/Introduction.tex
\section{Introduction}
Industrial Anomaly Detection (IAD) has significantly enhanced production efficiency and product quality by automatically identifying product defects~\cite{industrial_yingyong,noisyad,replaycad,3dad,miao2025robust,ijcai2025p80,ijcai2025p129}. However, traditional IAD methods primarily rely on single-view images, which inevitably suffer from blind spots and limited surface coverage.  To ensure comprehensive inspection of complex industrial components, the task of Multi-view Anomaly Detection (MAD) has emerged~\cite{real-iad,Manta}, aiming to utilize information from multiple viewpoints for a holistic evaluation.

\begin{figure*}[t]
\centerline{\epsfig{figure=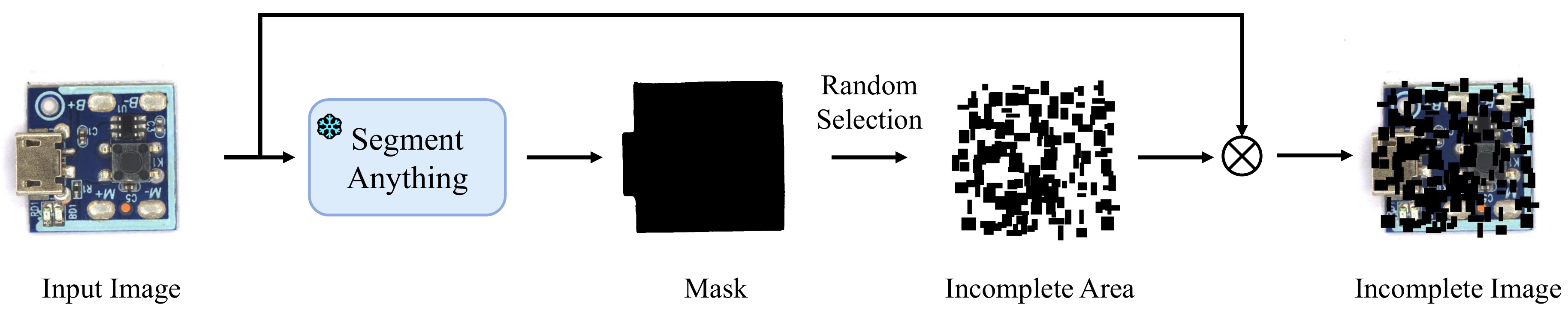,width=12cm}}
\caption{The overall production process of RIMAD. We first employ Segment Anything to extract the product masks, after which we simulate view incompleteness by randomly discarding image content within selected regions.}
\label{fig:2}
\end{figure*}

However, current mainstream MAD methods~\cite{mambaad,dinomaly,inp-former} usually assume that every view is complete and model each view separately. Essentially, this approach is still just single-view anomaly detection, and it fails to bring out the true potential of multi-view datasets. In real-world industrial scenarios, problems like occlusion or equipment errors often lead to missing information in different views. As a result, the performance of existing methods drops sharply because they lack the necessary correlation between these views, as shown in Figure \ref{fig:1}.

To tackle these limitations, we propose a more challenging task:  \textbf{I}ncomplete  \textbf{M}ulti-\textbf{V}iew  \textbf{A}nomaly  \textbf{D}etection ( \textbf{IMVAD}). In this task, parts of each view are masked, and the central challenge is to leverage information from other views to reconstruct the missing regions within the current view. To support this task, we proposed an automatic pipeline for generating the IMVAD dataset, as shown in Figure \ref{fig:2}. Specifically, we first employ Segment Anything (SAM)~\cite{SAM} to extract the product masks, after which we simulate view incompleteness by randomly discarding image content within selected regions. Following this procedure, we constructed the \textbf{RIMAD} dataset based on Real-IAD~\cite{real-iad}. In this dataset, 50\% of the target areas in each view are masked to reflect missing information. 

As illustrated in Figure~\ref{fig:1} and Table~\ref{tab:1}, we evaluated the performance of current state-of-the-art methods on the RIMAD dataset. Our observations reveal a substantial degradation in their performance, which can be attributed to two primary factors: (1) \textbf{The absence of cross-view consistency}, existing methods often treat each view independently and lack a mechanism to ensure consistency between different views. When a view is incomplete, the model cannot borrow useful information from other views to fill in the gaps, leading to unreliable detection results. (2) \textbf{Overfitting of the incomplete  area}, Conventional models are prone to overfitting to the specific patterns of missing data. Instead of learning robust features, the network may treat the masked regions as inherent data characteristics. This leads to a reduction in generalization, where the model struggles to differentiate between genuine structural anomalies and artifacts caused by data incompleteness.

To tackle these limitations, as illustrated in Figure \ref{fig:3}, we introduce \textbf{IMMoE}, a framework comprising two synergistic modules: (1) Multi-View Expert Fusion (\textbf{MVEF}), which leverages a multi-view expert network to effectively aggregate cross-view information, thereby providing robust guidance for single-view reconstruction and ensuring inter-view consistency. (2) Local Anomaly Enhancement Encoder (\textbf{LAEE}), which mitigates overfitting to masked regions by employing a localized dropout mechanism on feature maps. This regularizes the model to learn more discriminative representations and enhances its sensitivity to genuine anomalies. Furthermore, to enable the model to perceive the varying learning difficulties across different areas, we introduce a \textbf{Area Adaptive Loss}. Specifically, we partition the image into three distinct categories: incomplete product regions, complete product regions, and non-product (background) areas. By assigning different gradient weights to these three zones, we allow the model to adaptively focus on the most challenging parts, ensuring a more effective optimization process based on the inherent difficulty of each region. Extensive experiments show that our model achieves the best performance on both RIMAD and Real-IAD~\cite{real-iad} datasets, outperforming other existing methods.

To sum up, our contributions are as follows:

\begin{itemize}

    \item We identify and define a realistic and challenging task: Incomplete Multi-View Anomaly Detection (\textbf{IMVAD}). To support this research, we develop an automated generation pipeline and construct \textbf{RIMAD}, the first benchmark dataset specifically designed to evaluate model robustness under view incompleteness scenarios.

    \item We propose \textbf{IMMoE}, a novel framework tailored for the IMVAD task. By leveraging cross-view correlations and region-adaptive learning strategies, it effectively restores multi-view consistency and prevents the model from overfitting to masked regions.

    \item Extensive experiments demonstrate that our method achieves state-of-the-art performance on both the constructed RIMAD and the public Real-IAD datasets, especially on RIMAD, we have increased the pixel-level and image-level metrics by 11.8\% and 2.8\%, respectively.
\end{itemize}

%% file: section/Related_Work.tex
\section{Related Work}

\subsection{Multi-view Anomaly Detection}
Existing MAD methods can be broadly categorized into two types: \textbf{single-view-based} approaches and \textbf{multi-view fusion-based} approaches.

\textbf{Single-view-based} methods treat each view as an independent input and model them separately. UniAD~\cite{uniad} proposes a unified Transformer-based framework to handle multiple categories within a single network. SimpleNet~\cite{simplenet} achieves efficient localization by injecting noise into pre-trained features for discriminative training. MambaAD~\cite{mambaad} leverages State Space Models to enhance computational efficiency and capture long-range dependencies. Dinomaly~\cite{dinomaly} advocates a minimalist philosophy, demonstrating that pre-trained features require only minimal adaptation. INP-Former~\cite{inp-former} focuses on extracting intrinsic normal prototypes directly from individual images as robust references.

\textbf{Multi-view fusion-based} approaches aim to integrate information across multiple perspectives. IDIF~\cite{multiview-IDIF} introduces an intra-view decoupling and inter-view fusion strategy to better capture cross-view correlations. MVAD~\cite{mvad} introduces the MVAS algorithm, which utilizes a window-based attention mechanism to aggregate features from the most semantically correlated regions across multiple views. However, despite the cross-view interactions incorporated by these methods, they remain susceptible to overfitting on the incomplete regions when facing the IMVAD task, ultimately leading to a significant performance degradation.

\subsection{Mixture of Experts}

The Mixture of Experts (MoE) paradigm has emerged as a powerful architecture for scaling model capacity while maintaining computational efficiency through conditional computing~\cite{shazeer2017outrageously,AnomalyMoE,PromptMoE,Adapted-moe,Moead}. The core mechanism relies on a gating network that dynamically routes inputs to a sparse subset of specialized experts~\cite{jacobs1991adaptive}. Recent advancements have significantly extended the stability and efficiency of this routing process, such as the single-expert routing in Switch Transformers~\cite{switch2022} and the balanced competitive learning in V-MoE~\cite{vmoe2021}. MoEAD~\cite{meng2024moead} utilizes a Mixture-of-Experts pool within a recursive Transformer block. By adaptively selecting class-specific experts via a gating mechanism, it achieves parameter-efficient multi-class anomaly detection, balancing model compact-ness with high discriminative performance across diverse industrial categories. PromptMoE~\cite{PromptMoE} addresses zero-shot anomaly detection by replacing monolithic prompt learning with a compositional approach. It utilizes a pool of expert prompts (semantic primitives) and a visually-guided Mixture-of-Experts (MoE) mechanism. An image-gated router dynamically aggregates these experts to generate instance-adaptive textual representations, achieving superior generalization across diverse, unseen anomaly categories. Unlike scaling-oriented MoE, we reframe IMMoE for multi-view fusion. It leverages specialized experts to aggregate complementary information from all perspectives into a unified global representation. This global context then interacts with individual views, enabling the model to effectively restore incomplete regions by borrowing knowledge from the integrated multi-view features.

%% file: section/IMVAD.tex
\section{IMVAD}

\subsection{Problem Definition}

In real-world industrial scenarios, information loss in product images is often inevitable due to equipment malfunctions or physical occlusions. To evaluate model robustness under such realistic constraints, we propose a challenging task: Incomplete Multi-View Anomaly Detection (IMVAD). Specifically, the input for IMVAD in each viewpoint $i$ consists of two components: the incomplete view image $x_i$ and its corresponding binary mask $m_i$, the mask $m_i$ explicitly indicates the missing regions, where $m_i(p)=0$ denotes an incomplete pixel and $m_i(p)=1$ represents a valid pixel at position $p$. The model is required to reconstruct the original normal image even when the input view information is incomplete.

\subsection{RIMAD Dataset}
As illustrated in Figure~\ref{fig:2}, to construct the RIMAD dataset, we simulate realistic view incompleteness through a structured masking pipeline.  We first employ Segment Anything (SAM)~\cite{SAM} to precisely extract the product masks, ensuring that the simulation is focused solely on the object of interest.  To simulate information loss, we randomly discard image content within the segmented product regions.  Specifically, we populate the product area with randomly distributed rectangles, each with an area ranging from 8 to 32 square pixels.  This process is iteratively applied until \textbf{50\%} of the total product area is obscured.  This granular and stochastic masking strategy creates a challenging environment that forces the model to rely on cross-view correlations to reconstruct the missing industrial details.

%% file: section/Methods.tex
\section{Methods}

\subsection{Overview}
\begin{figure*}[t]
\centerline{\epsfig{figure=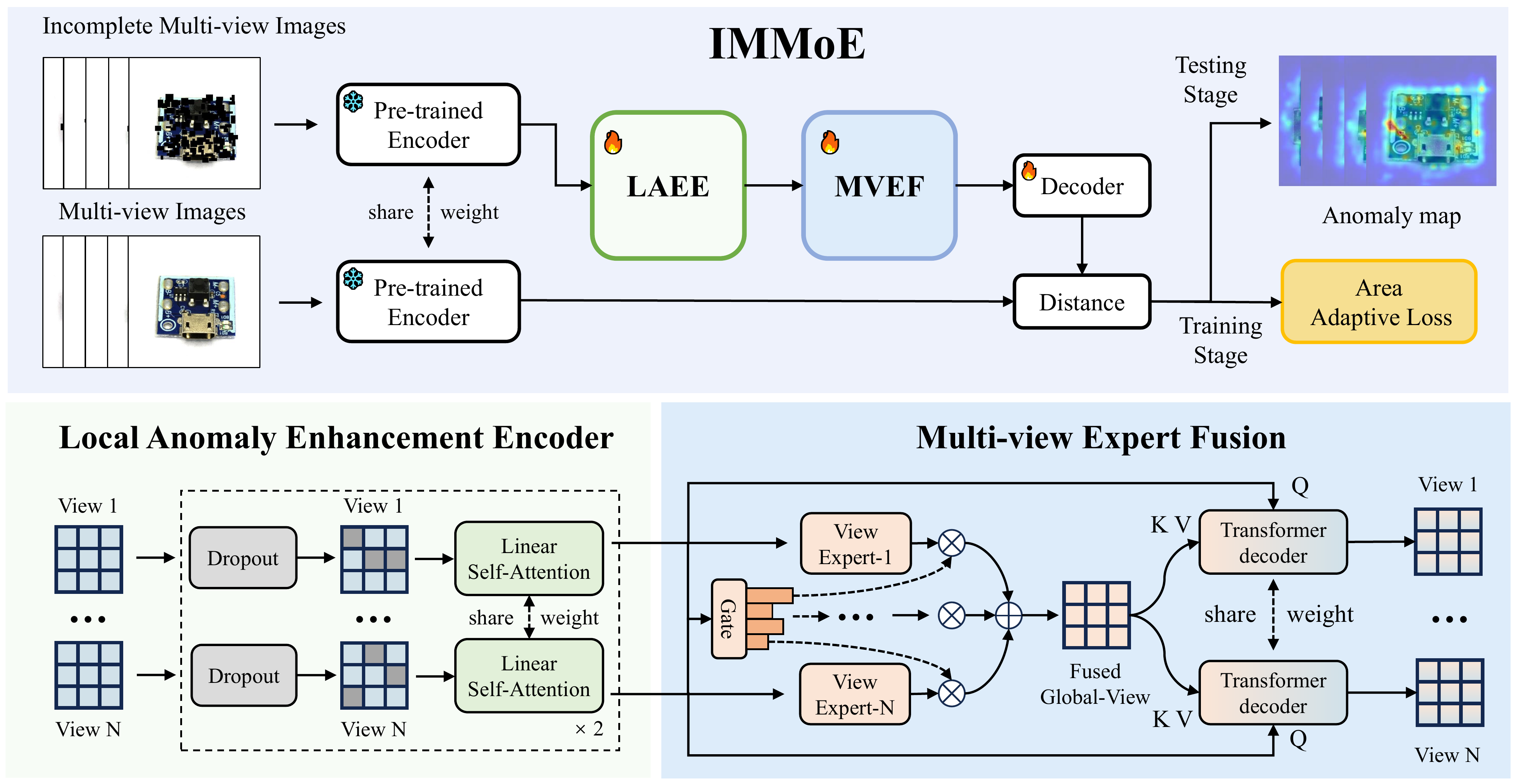,width=12cm}}
\caption{The overall architecture of the IMMoE framework. Our reconstruction-based pipeline centers on two synergistic modules: the \textbf{Local Anomaly Enhancement Encoder} (LAEE), which prevents overfitting to masked regions by simulating anomaly patterns through stochastic feature discarding, and the \textbf{Multi-view Expert Fusion} (MVEF), which adaptively aggregates cross-view complementary information via a Mixture-of-Experts mechanism. Together, they ensure robust feature restoration and inter-view consistency, enabling precise anomaly localization even under severe information deficiency.}
\label{fig:3}
\end{figure*}

The overall framework of our method is illustrated in Figure~\ref{fig:3}. Similar to INP-Former~\cite{inp-former} and Dinomaly~\cite{dinomaly}, we adopt a feature reconstruction-based framework. The overall framework consists of four key components and their corresponding intermediate features. First, in the \textbf{Feature Encoding} stage, a pre-trained encoder extracts representations from both the incomplete and original images, denoted as $F_{in}$ and $F_{gt}$, respectively. Next, $F_{in}$ is fed into the \textbf{Local Anomaly Enhancement Encoder} to generate enhanced features $F_{enh}$ by randomly discarding feature elements to simulate potential anomaly patterns. Subsequently, the \textbf{Multi-view Expert Fusion} module integrates these enhanced features across all perspectives, producing a fused representation $F_{fuse}$ that captures complementary information from multiple views. Finally, the \textbf{Reconstruction Decoder} utilizes $F_{fuse}$ to generate the restored features $F_{rec}$. During training, the model is optimized by minimizing the cosine distance between $F_{rec}$ and $F_{gt}$ using a area-adaptive loss, while in the inference phase, the anomaly map is derived directly from the discrepancy between these two feature sets.

\subsection{Local Anomaly Enhancement Encoder}

The primary motivation for the Local Anomaly Enhancement Encoder is to prevent the model from overfitting to the incomplete areas. Conventional reconstruction models are often prone to overfitting the specific patterns of missing data; instead of learning robust semantic features, the network may inadvertently treat the masked regions as inherent data characteristics. This leads to a significant reduction in generalization, where the model struggles to differentiate between genuine structural anomalies and artificial artifacts caused by data incompleteness. Inspired by Dinomaly~\cite{dinomaly} and Dropout~\cite{dropout}, we incorporate a stochastic feature-discarding mechanism to prevent the model from overfitting.

Specifically, for the incomplete view features $F_{in}^{i} \in {\mathbb{R}}^{N \times C}$, where $i$ denotes the $i$-th view of the sample, $N$ represents the number of image patches, and $C$ represents the feature dimension, we first employ a dropout mechanism to randomly discard feature elements:


\begin{eqnarray}
        F_{drop}^{i} = Dropout(F_{in}^{i},p)
        \text{,}
\end{eqnarray}

where $p$ denotes the dropout ratio.

Following the dropout operation, we incorporate a Linear Attention~\cite{dinomaly,linear_attention1,linear_attention2} mechanism to process the features of each view. The primary motivation is to recapture the spatial dependencies that may be disrupted by the stochastic discarding of features. By leveraging the global receptive field of attention, the model can effectively compensate for the missing information by aggregating context from the remaining active patches. We employ linear attention instead of standard self-attention to maintain computational efficiency while ensuring that each view's representation is sufficiently refined and robust before being passed to the subsequent fusion stage.

Specifically, the formula for Linear Attention (LA) is as follows:


\begin{eqnarray}
        Q = F_{drop}^{i} W^{Q},K = F_{drop}^{i} W^{K},V = F_{drop}^{i} W^{V}
        \text{,}
\end{eqnarray}


\begin{eqnarray}
        F_{enh}^{i} = LA(Q,K,V) = \sigma (Q)(\sigma(K^{T})V)
        \text{,}
\end{eqnarray}

where $W^{Q}, W^{K}, W^{V}$ are learnable parameters, $\sigma(x)$ is activation function.

Finally, the abnormally enhanced features of each view $F_{enh}^{i}$ will be further sent to the multi-view expert fusion network for global view fusion.

\subsection{Multi-view Expert Fusion}

The motivation for the Multi-view Expert Fusion (MVEF) module stems from the inherent information deficiency of individual views. While the LAEE enhances the robustness of features within a single perspective, it cannot inherently recover semantic information that is completely lost due to occlusion or masking. In industrial anomaly detection, different viewpoints often provide complementary geometric and structural information. For instance, a defect obscured in one view may be clearly visible or inferable from another. 

To fully exploit this multi-view synergy, the MVEF module is designed to dynamically aggregate cross-view knowledge. Instead of a simple concatenation or averaging, we employ a Mixture-of-Experts (MoE) architecture. This approach allows the model to adaptively select and weight information from different views based on their relevance to the missing regions. By establishing a global context through these specialized experts, the MVEF provides the necessary cross-view guidance for each individual view, ensuring that the subsequent reconstruction is informed by a comprehensive understanding of the object’s overall geometry rather than isolated, incomplete data.

Specifically, for the set of enhanced multi-view features $\{F_{enh}^i\}_{i=1}^V$ belonging to a single sample, we first concatenate them along the view dimension to serve as the input for the gating network:

\begin{eqnarray}
        F_{enh} = cat(F_{enh}^1;F_{enh}^2;\cdots;F_{enh}^V) \in {\mathbb{R}}^{V \times N \times C}
        \text{,}
\end{eqnarray}

where $V$ represents the total number of views.

The gating network is composed of a linear layer and a softmax function, which are utilized to calculate the importance scores for each individual view:

\begin{eqnarray}
        F_{gate} = softmax(linear(F_{enh})) \in {\mathbb{R}}^{V \times N}
        \text{,}
\end{eqnarray}

After obtaining the importance scores for each view, we employ a dedicated view-specific expert to encode each view independently. Each expert is constructed with two linear layers and an activation function, designed to extract deep semantic features from its corresponding perspective:

\begin{eqnarray}
        F_{expert}^i = linear_1^i(gelu(linear_2^i(F_{enh}^i))) \in {\mathbb{R}}^{N \times C}
        \text{,}
\end{eqnarray}

After obtaining the output from each view-specific expert, we compute the weighted sum by multiplying each expert's feature map by its corresponding importance score. These weighted features are then aggregated to produce the final global fused representation $ F_{fuse}$:

\begin{eqnarray}
        F_{fuse} = \textstyle\sum_{i=1}^{V}  F_{expert}^i F_{gate}^i, F_{fuse} \in {\mathbb{R}}^{N \times C}
        \text{,}
\end{eqnarray}

Upon obtaining the global fused representation $ F_{fuse}$, we employ a cross-attention~\cite{transformer} to integrate this global information back into each individual view:

\begin{eqnarray}
        Q = F_{enh}^{i} W^{Q},K = F_{fuse} W^{K},V = F_{fuse} W^{V}
        \text{,}
\end{eqnarray}

\begin{eqnarray}
        F_{enh}^{i} =  \text{softmax} \left( \frac{QK^T}{\sqrt{d_k}} \right) V
        \text{,}
\end{eqnarray}

Following the integration of global information, we employ a transformer-based~\cite{transformer,hu2024enhancing,lin2024handwriting,peng2025globally,peng2025proxy,lin2026atcmd} decoder utilizing linear attention to generate the final reconstructed features $F_{rec}^i$.

\subsection{Area Adaptive Loss}

To address the imbalance in reconstruction difficulty, inspired by INP-Former~\cite{inp-former} and Dinomaly~\cite{dinomaly}, we propose the Area Adaptive Loss (AAL), which dynamically modulates gradient magnitudes to prioritize informative regions over simple background textures.  Specifically, we scale the loss by the ratio of local loss to total batch loss and incorporate region-specific priors that assign higher weights to incomplete areas and product foregrounds.  This strategy ensures the optimization focuses on task-relevant and information-deficient regions, preventing the model from being overwhelmed by trivial background details. 

To minimize the discrepancy between the original and reconstructed features, we first define the \textbf{cosine distance} as:
\begin{equation}
    D_{cos}(a, b) = 1 - \frac{a^T \cdot b}{\|a\| \|b\|}
\end{equation}
The basic reconstruction loss for the $i$-th view is thus $\mathcal{L} = D_{cos}(F_{gt}^i, F_{rec}^i)$. We modulate the gradient of each patch $j$ by an adaptive weight $w(j)$, formulated as:
\begin{equation}
    \hat{f}(j) = cg(f(j)) \cdot w(j), \quad j \in [1, N]
\end{equation}
The weighting factor is defined as:

\begin{equation}
w(j) = ( \frac{D_{cos}(F_{gt}^{i,j}, F_{rec}^{i,j})}{\mu(D_{cos})} )^\theta + \beta
\end{equation}

where the first term emphasizes challenging patches by calculating the ratio of local loss to the batch mean $\mu(D_{cos})$. The second term, $\beta$, serves as a region-prior to incorporate spatial importance: we assign $\beta = 0.4$ for incomplete (masked) regions, $\beta = 0.2$ for product foregrounds, and $0$ otherwise. This dual-weighting mechanism ensures that the optimization prioritizes task-relevant and information-deficient areas over trivial background textures.

\begin{table*}[t]
\begin{center}
\scriptsize
\tabcolsep=0.1cm
\renewcommand\arraystretch{1.1}
\caption{Comparison with state-of-the-art methods on \textbf{RIMAD}. The results are presented in Image-AUROC/Pixel-AP. \textbf{Bold} indicates the best performance, while \underline{underline} indicates the second-best performance. All methods are reproduced under a unified experimental setting for fair comparison. ${\dagger}$: method designed for Multi-view Anomaly Detection. More results are presented in the supplementary materials.}
\begin{tabular}{ccccccc}
\specialrule{1.5pt}{0pt}{0pt} 
\rule{0pt}{2.5ex} Method → &  MambaAD  &  ViTAD  &  Dinomaly  &  INP-Former  &  MVAD$^{\dagger}$  &  \multirow{2}{*}{\textbf{Ours$^{\dagger}$}} \\ 
\cline{1-1}
 \rule{0pt}{2.5ex} Category ↓ &  NeurIPS 24  &  CVIU 25  &  CVPR 25  &   CVPR 25   &  TMM 26 \\ 
\hline
\rule{0pt}{2.5ex} Audiojack  &  81.7/32.8 &  78.0/21.5  &  \underline{83.1}/\underline{34.5}  &  77.1/27.8  &  76.7/26.9  &  \textbf{86.2}/\textbf{38.3} \\ 
Bottle Cap   &  \textbf{92.4}/\textbf{37.1} &  83.4/23.4  &  82.5/35.5  &  84.6/30.1  &  \underline{92.1}/\underline{36.8}  &  86.7/34.0 \\ 
Button Battery   &  64.8/20.7 &   65.6/39.6 &  74.0/\underline{39.7}  &  \underline{74.9}/\textbf{47.4}  &  62.3/23.4  &  \textbf{82.4}/32.9 \\ 
End Cap  &  \underline{73.0}/9.0 &  68.7/3.4  &  71.8/\underline{9.7}  &  67.9/4.8  &  71.9/8.0  &  \textbf{74.4}/\textbf{10.2} \\ 
Eraser  &  \textbf{89.0}/\underline{33.0} &  \underline{88.9}/30.6  &  83.9/30.3  &  88.3/\textbf{35.9}  &  88.8/32.5  &  85.0/30.4 \\ 
Fire Hood  &  82.7/\underline{34.4} &  77.8/24.3  &  71.0/13.1  &  \textbf{85.4}/\textbf{35.0}  &  79.9/33.7  &  \underline{83.4}/27.6 \\ 
Mint   &  65.3/8.7 &  68.6/\underline{12.7}  &  61.5/4.7  &  \textbf{71.8}/\textbf{13.7}  &  66.4/7.8  &  \underline{71.4}/11.9 \\ 
Mounts   &  \textbf{82.9}/\textbf{33.1} &  81.4/25.5  &  76.8/23.3  &  79.0/23.6  &  \underline{82.6}/\underline{33.0}  &  78.7/23.4 \\ 
PCB   &  \underline{81.7}/\underline{28.8} &  64.7/11.7  &  62.7/11.8  &  70.8/13.1  &  70.0/18.3  &  \textbf{86.1}/\textbf{45.2} \\ 
Phone Battery  &  85.4/20.8 &  78.8/16.4  &  \underline{85.6}/\underline{50.1}  &  84.8/31.4  &  79.9/17.9  &  \textbf{87.8}/\textbf{58.8} \\ 
Plastic Nut   &  86.5/\underline{34.4} &  81.4/21.6  &  77.3/12.9  &  \textbf{88.4}/\textbf{36.0}  &  85.8/32.9  &  \underline{88.1}/33.6 \\ 
Plastic Plug   &  79.2/\textbf{28.2} &  71.4/11.0  &  47.5/1.9  &  \textbf{83.8}/25.0  &  77.8/\underline{26.8}  &  \underline{80.8}/20.7 \\ 
Porcelain Doll   &  83.3/\underline{27.5} &  80.2/22.2  &  82.8/18.2  &  \textbf{88.7}/24.6  &  84.1/\textbf{28.2}   &  \underline{85.1}/19.2 \\ 
Regulator   &  43.8/0.5 &  45.2/0.4  &  \underline{60.8}/\underline{9.6}  &  60.6/5.1  &  42.4/0.3  &  \textbf{70.6}/\textbf{20.4} \\ 
Rolled Strip Base  &  92.6/30.3 &  \underline{96.1}/29.0  &  94.6/\underline{48.6}  &  93.1/36.7  &  93.3/29.7  &  \textbf{96.4}/\textbf{49.6} \\ 
SIM Card Set   &  92.6/41.8 &  92.7/40.2  &  94.4/\underline{44.8}  &  \underline{94.9}/44.6  &  87.6/\textbf{49.3}   &  \textbf{95.2}/44.4 \\ 
Switch  &  \underline{85.4}/34.8 &  78.4/32.5  &  77.0/26.7  &  82.9/\underline{35.5}  &  78.7/22.7  &  \textbf{94.4}/\textbf{59.4} \\ 
Tape   &  \textbf{95.6}/\textbf{43.1} &  93.2/34.2  &  89.5/\textbf{43.1}  &  93.7/42.6  &  \underline{95.0}/\underline{43.0}  &  91.5/42.6 \\ 
Terminal Block  &  \textbf{93.9}/40.1 &  83.4/21.5  &  90.1/\underline{48.0}  &  88.8/40.5  &  90.1/36.3  &  \underline{92.6}/\textbf{51.0} \\ 
Toothbrush  &  \textbf{83.2}/\underline{27.8} &  \underline{82.9}/\textbf{30.1}  &  70.0/12.6  &  80.8/19.4  &  79.3/24.3  &   82.1/22.0 \\ 
Toy  &  63.9/5.4 &  66.7/7.0  &  70.2/\underline{9.5}  &  \underline{70.3}/9.2  &  56.2/4.8  &  \textbf{74.0}/\textbf{12.3} \\ 
Toy Brick  &  \underline{81.9}/\textbf{27.2} &  75.3/23.9  &  70.4/14.6  &  74.7/17.9  &  \textbf{82.0}/\underline{26.6}  &  72.2/14.2 \\ 
Transistor1  &  \textbf{92.0}/37.3 &  84.1/32.3  &  87.9/\textbf{44.9}  &  88.1/\underline{42.9}  &  87.1/32.6  &  \underline{89.3}/\textbf{44.9} \\ 
U Block  &  77.0/24.6 &  77.6/\textbf{28.0}  &  82.1/20.9  &  \underline{82.6}/\underline{26.1}  &  77.0/23.6  &  \textbf{85.0}/25.8 \\ 
USB  &  82.4/36.5 &  77.2/21.1  &  \underline{87.2}/\underline{39.1}  &  81.6/21.7  &  69.5/21.9  &  \textbf{91.6}/\textbf{42.5} \\ 
USB Adaptor  &  82.4/\underline{24.3} &  76.0/13.3  &  81.3/\textbf{24.8}  &  \textbf{83.4}/21.5  &  80.9/23.1  &  \underline{82.7}/23.0 \\ 
Vcpill   &  79.6/43.0 &  85.2/\textbf{55.5}  &  65.4/21.2  &  \textbf{89.7}/\underline{53.8}  &  79.4/39.8  &  \underline{87.0}/50.1 \\ 
Wooden Beads   &  \textbf{85.6}/30.0 &  80.0/\underline{31.4}  &  80.6/26.8  &  84.7/\textbf{32.4}  &  \underline{84.9}/30.1  &  82.9/26.6 \\ 
Woodstick  &  79.0/\underline{40.4} &  72.1/37.3  &  76.7/36.9  &  \underline{79.4}/\textbf{42.3}  &  79.1/40.1  &  \textbf{80.1}/34.3 \\ 
Zipper  &  97.8/32.2 &  97.6/\textbf{56.0}  &  92.8/\underline{54.9}  &  \underline{98.3}/47.1  &  97.6/48.4  &  \textbf{98.6}/45.1 \\ 
\hline \rowcolor{cyan!8}
\rule{0pt}{2.5ex} \textbf{Average}   &  81.9/28.9 &  78.4/25.2  &  77.7/27.1  &  \underline{82.4}/\underline{29.6}  &  79.3/27.4  &  \textbf{84.7}/\textbf{33.1} \\ 
\specialrule{1.5pt}{0pt}{0pt}
\end{tabular}
\label{tab:1}
\end{center}
\end{table*}

%% file: section/Experiments.tex
\begin{table}[t]
\begin{center}
\tabcolsep=0.3cm
\renewcommand\arraystretch{1}
\caption{Comparison with state-of-the-art methods on \textbf{Real-IAD}. \textbf{Bold} indicates the best performance, while \underline{underline} indicates the second-best performance. ${\dagger}$: method designed for Multi-view Anomaly Detection. All methods are reproduced under a unified experimental setting for fair comparison.}
\begin{tabular}{cccc}
\specialrule{1.5pt}{0pt}{0pt}
\textbf{Method} & \textbf{Venue} & I-AUROC & P-AP \\ \hline
 UniAD & NeurIPS 22 & 83.0 & 21.1 \\
  DeSTSeg  & CVPR 23  &  82.3  & 37.9  \\
 SimpleNet & CVPR 23 & 71.1 & 11.5 \\
 DiAD &  AAAI 24  &  75.6  & 2.9  \\
 MambaAD & NeurIPS 24  & 86.3 & 33.0 \\
 ViTAD& CVIU 25 & 82.7 & 24.3 \\
 Dinomaly & CVPR 25 &  \underline{89.3} & 42.8 \\
INP-Former & CVPR 25 & 89.1 &  \underline{43.7} \\
 MVAD$^{\dagger}$ & TMM 26 & 86.6 & 30.3 \\ \hline \rowcolor{cyan!8}
 \multicolumn{2}{c}{\textbf{Ours$^{\dagger}$}}  & \textbf{89.7} & \textbf{44.3} \\ \specialrule{1.5pt}{0pt}{0pt}
\end{tabular}
\label{tab:2}
\end{center}
\end{table}

\section{Experiments}

\subsection{Experiments Setup}

{\bf Datasets.} We conducted our experiments on the two datasets, RIMAD and Real-IAD~\cite{real-iad}. The Real-IAD contains samples of a total of 30 categories, and each sample contains images of 5 views. There were a total of 99,721 normal sample images, of which 36,465 were used for training and the remaining 63,256 for testing. There were 51,329 abnormal sample images, all of which were used for testing. We obtained the RIMAD dataset by masking 50\% of the area in each view of the Real-IAD dataset. The specific method is shown in Figure~\ref{fig:2}.

{\bf Compared Methods.} We compared with the most advanced multi-view anomaly detection methods currently available, including: UniAD~\cite{uniad}, DeSTSeg~\cite{destseg}, SimpleNet~\cite{simplenet}, DiAD~\cite{diad}, MambaAD~\cite{mambaad}, ViTAD~\cite{vitad}, Dinomaly~\cite{dinomaly}, INP-Former~\cite{inp-former}, MVAD~\cite{mvad}.

{\bf Metrics.} Consistent with previous methods, we evaluate image-level anomaly classification using Area Under the Receiver Operating Characteristics (AUROC)~\cite{AUROC}. For pixel-level anomaly segmentation, we adopt Area Under Precision-Recall (AP)~\cite{AP}. We prioritize AP for segmentation because anomaly regions are significantly smaller than normal areas, AP provides a more reliable assessment by focusing on the minority class in such highly imbalanced pixel distributions.

{\bf Implementation Details.} We adopt ViT-Base/14 with DINO2-R~\cite{dino} as the pre-trained image encoder. The input image is first resized to $448^{2}$ and then center-cropped to $392^{2}$. We employ the StableAdamW optimizer~\cite{StableAdamW} equipped with AMSGrad~\cite{AMSGrad} to train the network for 50,000 iterations. The mask ratio of the RIMAD dataset has been set to 50\%. We set the learning rate to $2\text{e-}3$, $\beta$ coefficients to $(0.9, 0.999)$, and weight decay to $1\text{e-}4$. We set the dropout ratio $p$ to 0.1, and $\theta$ to 3. All the experiments were conducted on a single NVIDIA RTX 4090 GPU.

\begin{figure*}[t]
\centerline{\epsfig{figure=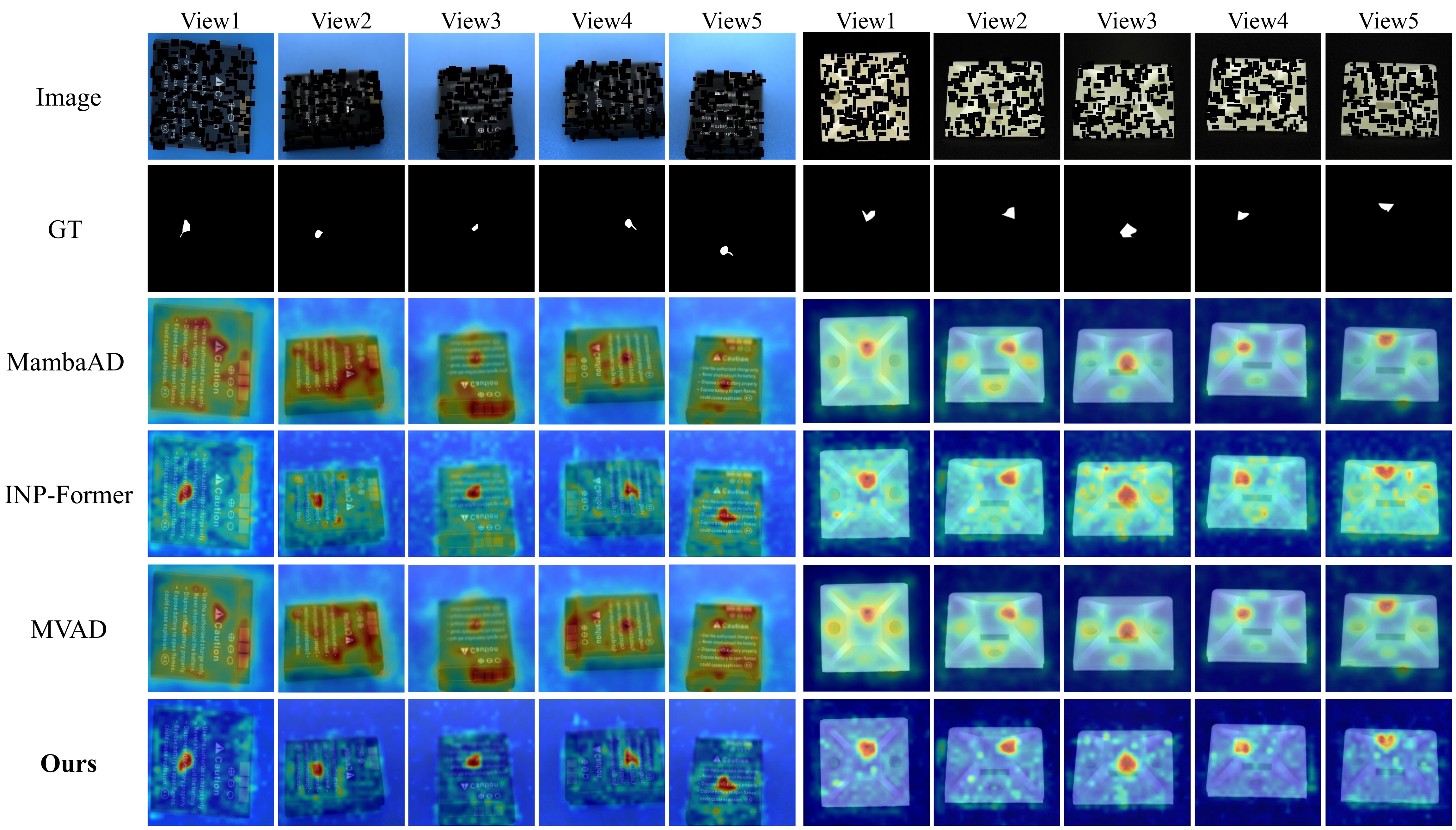,width=12cm}}
\caption{Visualization of anomalous region segmentation. Areas with high anomaly scores are shown in red, while areas with low anomaly scores are shown in blue.}
\label{fig:4}
\end{figure*}

\subsection{Comparison with Sota Methods}

{\bf Quantitative Evaluation.} As demonstrated in Table~\ref{tab:1} and Table~\ref{tab:2}, our IMMoE framework achieves state-of-the-art (SOTA) performance across both RIMAD and Real-IAD datasets.

On the RIMAD dataset (Table~\ref{tab:1}), which specifically simulates view incompleteness, our method reaches 84.7\% I-AUROC and 33.1\% P-AP, outperforming the SOTA single-view method INP-Former~\cite{inp-former} by 2.8\% and 11.8\%, respectively. Crucially, compared to the specialized multi-view method MVAD~\cite{mvad}, our approach shows a significant margin of 5.4\% in I-AUROC and 5.7\% in P-AP. This gap reveals that while MVAD struggles to aggregate reliable features from corrupted inputs (79.3\% I-AUROC), our MVEF module effectively reconstructs missing semantic priors by synergizing information across viewpoints.

On the Real-IAD dataset (Table~\ref{tab:2}), where all five views are complete, our method maintains its lead with 89.7\% I-AUROC and 44.3\% P-AP. It is noteworthy that while many recent high-performing methods like Dinomaly~\cite{dinomaly} and INP-Former~\cite{inp-former} show strong results on complete data, our framework still surpasses them, further widening the gap over MVAD~\cite{mvad} (86.6\% I-AUROC and 30.3\% P-AP). This consistent superiority across both datasets underscores that our view-expert fusion strategy is not merely a remedy for missing data, but a robust architecture capable of adaptively capturing complex cross-view correlations regardless of information density.

{\bf Qualitative Evaluation.} Figure~\ref{fig:4} presents the heatmap visualizations of the anomaly regions predicted by our proposed method alongside other state-of-the-art techniques. It is evident that existing methods frequently struggle with information deficiency in incomplete views, often resulting in large-scale prediction errors, such as misidentifying normal background textures as anomalies or producing fragmented, noisy localization maps. In sharp contrast, by effectively integrating synergistic multi-view information, our approach suppresses such background interference and yields significantly more precise and comprehensive anomaly localizations. Our heatmaps demonstrate superior spatial consistency and align more closely with the actual defect boundaries, confirming that our model's cross-view reconstruction capability is crucial for high-fidelity anomaly segmentation even under severe occlusions.

\begin{table}[t]
\begin{center}
\tabcolsep=0.2cm
\renewcommand\arraystretch{1}
\caption{ Ablations of model design on RIMAD. LAEE: Local Anomaly Enhancement Encoder. MVEF: Multi-view Expert Fusion. AAL: Area
Adaptive Loss.}
\begin{tabular}{ccc|cc}
\specialrule{1.5pt}{0pt}{0pt}
 LAEE & MVEF & AAL & I-AUROC & P-AP \\ \hline
 \ding{55} & \ding{55} & \ding{55} &76.9&26.2  \\
\ding{51} & \ding{55} & \ding{55} &  81.3 & 29.1  \\
 \ding{51} & \ding{51} & \ding{55} & \underline{84.0} &  \underline{32.3} \\ 
 \ding{51} & \ding{51} & \ding{51} &{\bf84.7}&{\bf33.1} \\ \specialrule{1.5pt}{0pt}{0pt}
\end{tabular}
\label{tab:3}
\end{center}
\end{table}

\begin{table}[t]
\begin{center}
\tabcolsep=0.3cm
\renewcommand\arraystretch{1}
\caption{ Ablations of mask ratio on RIMAD (default=50\%).}
\begin{tabular}{c|ccc}
\specialrule{1.5pt}{0pt}{0pt}
Ratio & Dinomaly & INP-Former & Ours \\ \hline
0\%& \underline{89.3} / 42.8 &89.1 / \underline{43.7}& {\bf89.7} / {\bf44.3}   \\
25\%& 76.5 / \underline{21.6} & \underline{77.1} /  {\bf22.1}&  {\bf77.9} / 20.0 \\
50\%& 77.7 / 27.1 &\underline{82.4} / \underline{29.6}& {\bf84.7} / {\bf33.1}  \\
75\%& \underline{84.4} / \underline{39.8} & 83.7 / 34.1& {\bf87.3} / {\bf42.0} \\ \specialrule{1.5pt}{0pt}{0pt}
\end{tabular}
\label{tab:4}
\end{center}
\end{table}

\begin{table}[t]
\begin{center}
\tabcolsep=0.3cm
\renewcommand\arraystretch{1}
\caption{ Ablations of dropout rate on RIMAD (default=0.1).}
\begin{tabular}{c|cc}
\specialrule{1.5pt}{0pt}{0pt}
 Dropout rate & I-AUROC & P-AP \\ \hline
 0.0  &  83.0 &  29.6   \\
 0.1 & {\bf84.7} & \underline{33.1} \\
0.2  & \underline{84.2} & {\bf33.4}  \\
0.3  & 76.9 & 26.8 \\ \specialrule{1.5pt}{0pt}{0pt}
\end{tabular}
\label{tab:5}
\end{center}
\end{table}

\subsection{Ablation Study}

{\bf Model Design.} To evaluate the contribution of each component, we conduct an ablation study on the RIMAD dataset (Table~\ref{tab:3}). Integrating the Local Anomaly Enhancement Encoder (LAEE) raises the Image-AUROC from 76.9\% to 81.3\%, confirming its ability to prevent overfitting through feature dropout. The addition of Multi-view Expert Fusion (MVEF) further boosts performance to 84.0\% I-AUROC, demonstrating that cross-view synergy is vital for recovering lost semantic details. Finally, the Area Adaptive Loss (AAL) yields the best results (84.7\% I-AUROC, 33.1\% P-AP), proving that dynamic gradient modulation effectively prioritizes challenging, task-relevant regions. These consistent gains validate the complementary nature of our synergistic modules.

\begin{figure*}[t]
\centerline{\epsfig{figure=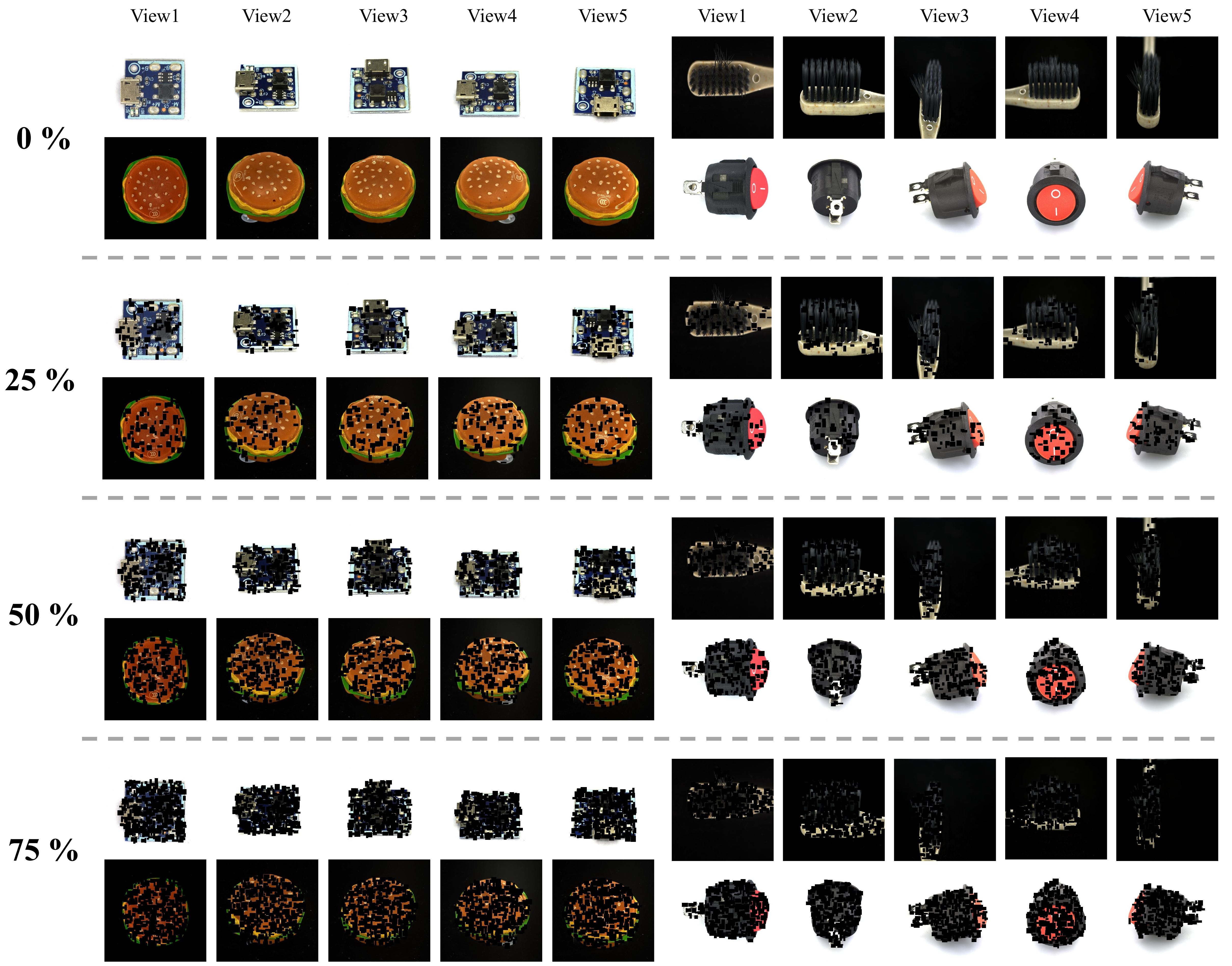,width=12cm}}
\caption{Visualization of different mask ratio on RIMAD (default=50\%).}
\label{fig:5}
\end{figure*}

{\bf Mask Ratio.} As illustrated in Table ~\ref{tab:4} and Figure~\ref{fig:5}, we investigate the model's sensitivity to different mask ratios on RIMAD. An intriguing observation is that performance is relatively lower at moderate mask ratios (e.g., 25\%) but improves as the ratio increases to 75\%. We attribute this to a shift in the model's learning objective: at lower mask ratios, the model tends to rely heavily on the remaining local spatial features of the current view, which may be ambiguous or contain misleading cues from the incomplete boundaries. In contrast, at higher mask ratios, the model is "forced" to abandon local dependencies and instead functions as a pure reconstruction task, leveraging robust cross-view semantic priors and global context. This transition reduces the interference from corrupted local information, allowing the Multi-view Expert Fusion (MVEF) to more effectively aggregate reliable information from other views, thereby leading to superior and more stable anomaly detection performance.

{\bf Dropout Rate.} We investigate the sensitivity of the LAEE to the dropout rate $p$, as shown in Table~\ref{tab:5}. The performance peaks at a dropout rate of 0.1, demonstrating that a moderate level of feature discarding effectively regularizes the model and prevents overfitting to masked regions. When the dropout rate is too low (0.0), the model lacks sufficient regularization, whereas a rate that is too high (0.3) leads to an excessive loss of critical semantic information, causing a sharp drop in performance. This parabolic trend suggests that an optimal dropout rate is essential for balancing feature robustness with information integrity, ensuring the model focuses on learning generalizable anomaly patterns.

%% file: section/sup.tex
\clearpage

\appendix
\section{Appendix}

This supplementary material consists of:

\begin{itemize}

   \item In Section \ref{sec:a1}, we have provided more ablation experiments.

  \item In Section \ref{sec:a2}, we have provided more results on RIMAD.

  \item  In Section \ref{sec:a3}, we provide image-level and pixel-level evaluation metrics for each class on Real-IAD.

    \item  In Section \ref{sec:a4}, we provide more qualitative evaluation results on RIMAD.

\end{itemize}
\subsection{More Ablation Experiments}
\label{sec:a1}
As illustrated in Table~\ref{tab:a1}, we investigated the impact of different numbers of views, thereby simulating a situation where one or more views might be completely missing in a real industrial scenario. The experimental results clearly demonstrate a positive correlation between the anomaly detection performance and the number of available input views. When the system degrades to an extreme single-view input (View=1), the Image-AUROC and Pixel-AP drop to 80.3\% and 29.5\%, respectively, primarily due to severe spatial information loss and the self-occlusion of the objects. As the number of viewpoints progressively increases, both metrics exhibit a steady upward trend. Notably, a significant performance leap occurs when the view count increases from 3 to 4, yielding a 1.8\% improvement in I-AUROC and a 1.5\% increase in P-AP. This indicates that the introduction of multiple perspectives effectively supplements the side and edge details of the targets, successfully overcoming the spatial blind spots inherent in single-view observations. Ultimately, the model reaches its peak performance (84.7\% / 33.1\%) under the default full 5-view setting. Furthermore, it is worth emphasizing that even under the stringent condition of randomly losing two views (View=3), our approach still maintains a competitive I-AUROC of 82.2\%. This strongly validates that our proposed feature fusion mechanism not only maximizes the utilization of comprehensive spatial cues but also possesses remarkable robustness and fault tolerance against unexpected situations common in actual industrial production lines, such as camera malfunctions or image transmission failures.

\begin{table}[]
\begin{center}
\tabcolsep=0.2cm
\renewcommand\arraystretch{1.1}
\caption{ Ablations of different input view numbers on RIMAD (default=5).}
\begin{tabular}{c|cc}
\specialrule{1.5pt}{0pt}{0pt}
 View 
 numbers & I-AUROC & P-AP \\ \hline
 1  &  80.3  & 29.5    \\
 2 &  80.7 & 30.0  \\
3  & 82.2  &  31.4  \\
4  &  84.0 &  32.9  \\
5  & 84.7 & 33.1 \\ \specialrule{1.5pt}{0pt}{0pt}
\end{tabular}
\label{tab:a1}
\end{center}
\end{table}

\subsection{More Results on RIMAD }
\label{sec:a2}

As illustrated in Table~\ref{tab:a2}, we provide further comparison results on the RIMAD dataset by introducing three additional state-of-the-art baselines: RD~\cite{rd}, UniAD~\cite{uniad}, and SimpleNet~\cite{simplenet}. Evaluated under a unified setting, our method consistently outperforms these newly added approaches. Specifically, our model achieves the highest average Image-AUROC of \textbf{84.7\%} and Pixel-AP of \textbf{33.1\%} across all 30 object categories, demonstrating superior robustness and generalization capabilities for diverse industrial anomaly detection.

\subsection{Detailed results on Real-IAD}
\label{sec:a3}
Table \ref{tab:a3} provides a comprehensive per-category comparison on the Real-IAD~\cite{real-iad} dataset, encompassing 30 industrial categories. Our proposed method, specifically designed for multi-view scenarios (${\dagger}$), achieves the highest average performance with 89.7\% Image-AUROC and 44.4\% Pixel-AP, consistently outperforming recent state-of-the-art methods. A detailed category-wise analysis reveals several key insights. First, our method demonstrates exceptional robustness in complex geometries, such as Audiojack, Eraser, and Tape, where it secures the top position in both image-level detection and pixel-level localization. Second, while INP-Former~\cite{inp-former} and Dinomaly~\cite{dinomaly} show competitive results in specific categories like SIM Card Set and Switch, our approach maintains a more stable performance across the entire dataset, particularly in Pixel-AP where it leads the second-best method (INP-Former~\cite{inp-former}) by 0.7\%. Notably, compared to the dedicated multi-view baseline MVAD~\cite{mvad}, our model achieves a significant gain of 14.1\% in Pixel-AP.

\subsection{More Qualitative Evaluation}
\label{sec:a4}

\begin{figure*}[]
\centerline{\epsfig{figure=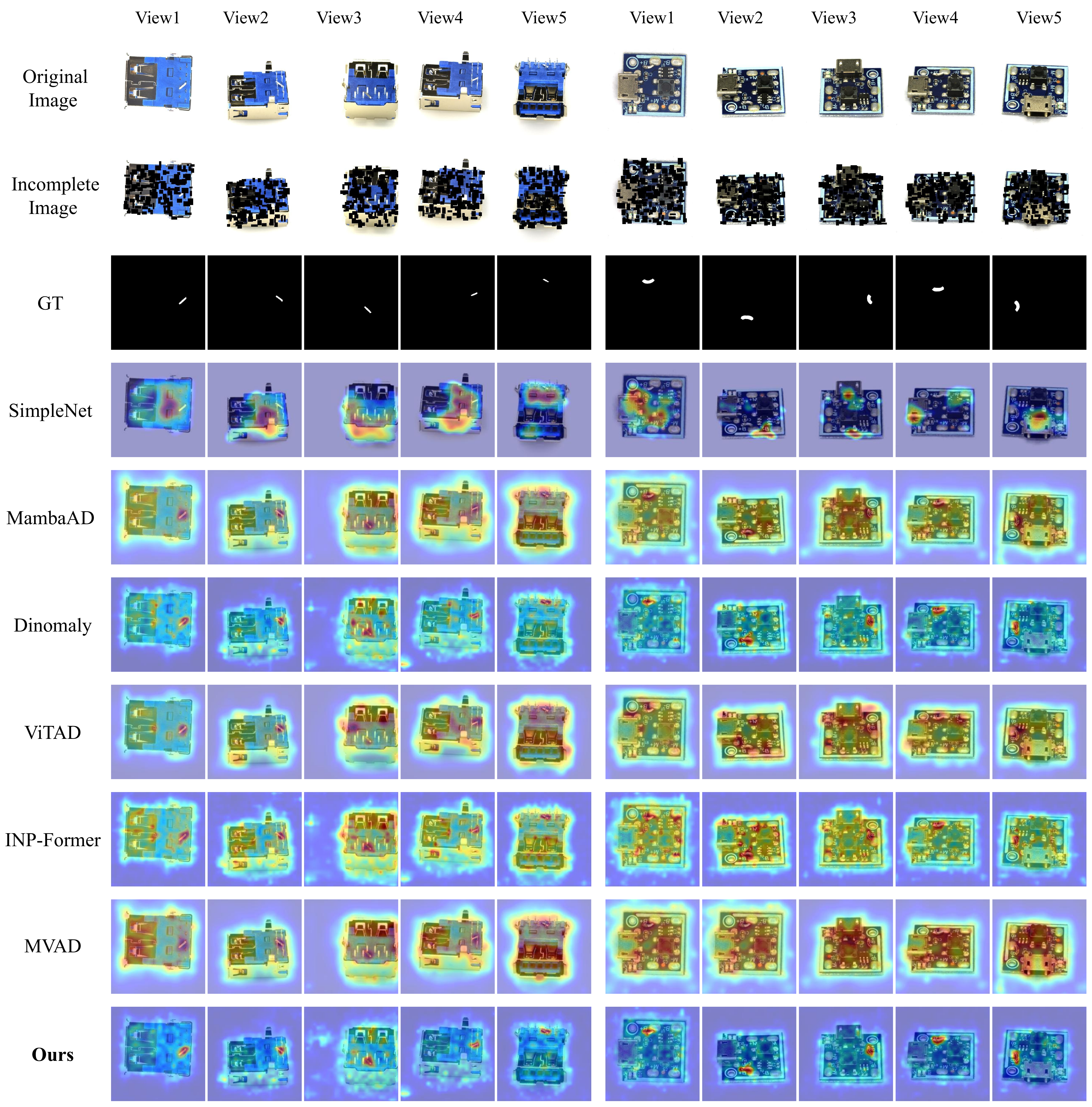,width=12cm}}
\caption{Visualization of anomalous region segmentation on RIMAD. Areas with high anomaly scores are shown in red, while areas with low anomaly scores are shown in blue.}
\label{fig:a1}
\end{figure*}

As illustrated in Figure~\ref{fig:a1}, we have provided more qualitative comparison results for the segmentation of abnormal areas. Including the following comparison methods: SimpleNet~\cite{simplenet}, MambaAD~\cite{mambaad}, ViTAD~\cite{vitad}, Dinomaly~\cite{dinomaly}, INP-Former~\cite{inp-former}, MVAD~\cite{mvad}. While competing methods frequently suffer from perspectival distortions and environmental noise—leading to fragmented heatmaps or significant false positives in normal regions—our approach consistently produces clean and semantically coherent anomaly maps. By effectively leveraging synergistic multi-view information, our model filters out complex background interference and achieves superior spatial consistency. Even for subtle defects and intricate geometries, our predicted regions align much more closely with the ground-truth boundaries. These results further demonstrate that our cross-view reconstruction framework is essential for achieving robust and high-fidelity anomaly localization in diverse industrial scenarios.

\begin{table}[]
\begin{center}
\tabcolsep=0.3cm
\renewcommand\arraystretch{1.1}
\caption{Comparison with more state-of-the-art methods on \textbf{RIMAD}. The results are presented in Image-AUROC/Pixel-AP. \textbf{Bold} indicates the best performance, while \underline{underline} indicates the second-best performance. All methods are reproduced under a unified experimental setting for fair comparison. ${\dagger}$: method designed for Multi-view Anomaly Detection.}
\begin{tabular}{ccccc}
\specialrule{1.5pt}{0pt}{0pt} 
\rule{0pt}{2.5ex} Method → & RD & UniAD & SimpleNet & \multirow{2}{*}{\textbf{Ours$^{\dagger}$}} \\ \cline{1-1}
\rule{0pt}{2.5ex} Category ↓ & CVPR 22 & NeurIPS 22 & CVPR 23 & \\ \hline
\rule{0pt}{2.5ex} Audiojack & \underline{74.1} / \underline{22.0} & \textbf{76.4} / 18.7 & 71.4 / 2.6 & \textbf{86.2} / \textbf{38.3} \\ 
Bottle Cap & \underline{85.7} / \underline{24.6} & 83.8 / 21.5 & 63.0 / 3.9 & \textbf{86.7} / \textbf{34.0} \\ 
Button Battery & \underline{69.6} / \textbf{43.9} & 60.4 / 10.8 & 55.6 / 13.3 & \textbf{82.4} / \underline{32.9} \\ 
End Cap & \textbf{72.9} / 1.4 & 70.3 / \underline{6.7} & 60.2 / 0.6 & \textbf{74.4} / \textbf{10.2} \\ 
Eraser & 81.9 / 22.7 & \textbf{88.8} / \underline{27.5} & 59.5 / 3.5 & \underline{85.0} / \textbf{30.4} \\ 
Fire Hood & 74.3 / \underline{24.1} & \underline{78.4} / 22.9 & 62.5 / 2.9 & \textbf{83.4} / \textbf{27.6} \\ 
Mint & 60.8 / 2.9 & \underline{64.5} / \underline{5.6} & 52.8 / 1.2 & \textbf{71.4} / \textbf{11.9} \\ 
Mounts & 75.2 / 21.7 & \textbf{84.2} / \textbf{28.5} & 72.3 / 4.2 & \underline{78.7} / \underline{23.4} \\ 
PCB & \underline{80.0} / \underline{27.9} & 72.6 / 4.3 & 60.9 / 4.3 & \textbf{86.1} / \textbf{45.2} \\ 
Phone Battery & \underline{80.9} / \underline{16.8} & 75.4 / 3.9 & 66.4 / 1.2 & \textbf{87.8} / \textbf{58.8} \\ 
Plastic Nut & \underline{80.5} / \underline{21.3} & 65.8 / 10.5 & 59.8 / 2.5 & \textbf{88.1} / \textbf{33.6} \\ 
Plastic Plug & \underline{73.6} / \underline{16.4} & 70.0 / 9.8 & 61.3 / 1.7 & \textbf{80.8} / \textbf{20.7} \\ 
Porcelain Doll & 75.9 / \underline{14.5} & \underline{79.6} / 11.4 & 71.8 / 6.5 & \textbf{85.1} / \textbf{19.2} \\ 
Regulator & \underline{58.5} / \underline{5.9} & 42.0 / 0.3 & 47.6 / 0.4 & \textbf{70.6} / \textbf{20.4} \\ 
Rolled Strip Base & 87.9 / \underline{26.7} & \underline{89.6} / 18.4 & 65.8 / 0.9 & \textbf{96.4} / \textbf{49.6} \\ 
SIM Card Set & \underline{85.6} / \underline{37.0} & 84.9 / 28.5 & 78.2 / 11.1 & \textbf{95.2} / \textbf{44.4} \\ 
Switch & \underline{83.1} / \underline{36.3} & 76.7 / 32.1 & 66.6 / 16.5 & \textbf{94.4} / \textbf{59.4} \\ 
Tape & 87.4 / 31.2 & \textbf{94.9} / \underline{34.7} & 65.2 / 6.4 & \underline{91.5} / \textbf{42.6} \\ 
Terminal Block & \underline{86.3} / \underline{29.5} & 82.3 / 19.9 & 63.7 / 1.8 & \textbf{92.6} / \textbf{51.0} \\ 
Toothbrush & \underline{77.8} / \underline{22.7} & 73.5 / 15.8 & 68.9 / 14.0 & \textbf{82.1} / \textbf{22.0} \\ 
Toy & \underline{69.0} / \underline{2.8} & 56.1 / 0.9 & 55.6 / 0.4 & \textbf{74.0} / \textbf{12.3} \\ 
Toy Brick & \underline{72.7} / \textbf{20.0} & \textbf{78.7} / 18.6 & 57.6 / 5.2 & 72.2 / \underline{14.2} \\ 
Transistor1 & 86.5 / \underline{29.5} & \textbf{90.9} / 24.1 & 75.4 / 11.2 & \underline{89.3} / \textbf{44.9} \\ 
U Block & \underline{78.4} / \underline{20.7} & \underline{80.5} / 19.4 & 66.3 / 4.1 & \textbf{85.0} / \textbf{25.8} \\ 
USB & \underline{82.8} / \underline{30.5} & 73.8 / 10.1 & 64.9 / 4.3 & \textbf{91.6} / \textbf{42.5} \\ 
USB Adaptor & \underline{75.5} / \underline{8.3} & 71.9 / 7.4 & 55.3 / 0.2 & \textbf{82.7} / \textbf{23.0} \\ 
Vcpill & 79.0 / \underline{42.1} & \underline{84.0} / \textbf{44.0} & 70.4 / 8.9 & \textbf{87.0} / \textbf{50.1} \\ 
Wooden Beads & \underline{75.5} / \underline{19.8} & \underline{76.5} / 17.8 & 62.4 / 8.1 & \textbf{82.9} / \textbf{26.6} \\ 
Woodstick & 72.1 / \underline{30.7} & \underline{77.4} / \textbf{36.0} & 66.9 / 18.1 & \textbf{80.1} / \underline{34.3} \\ 
Zipper & 88.3 / \underline{41.1} & \underline{95.7} / 26.3 & 88.2 / 25.8 & \textbf{98.6} / \textbf{45.1} \\ \hline \rowcolor{cyan!8}
\rule{0pt}{2.5ex} \textbf{Average} & \underline{77.7} / \underline{23.2} & 76.6 / 17.9 & 64.0 / 6.2 & \textbf{84.7} / \textbf{33.1} \\ 
\specialrule{1.5pt}{0pt}{0pt}
\end{tabular}
\label{tab:a2}
\end{center}
\end{table}

\begin{table}[]
\begin{center}
\scriptsize
\tabcolsep=0.12cm
\renewcommand\arraystretch{1.1}
\caption{Detailed results for each category  on \textbf{Real-IAD}. The results are presented in Image-AUROC/Pixel-AP. \textbf{Bold} indicates the best performance, while \underline{underline} indicates the second-best performance. All methods are reproduced under a unified experimental setting for fair comparison. ${\dagger}$: method designed for Multi-view Anomaly Detection.}
\begin{tabular}{ccccccc}
\specialrule{1.5pt}{0pt}{0pt} 
\rule{0pt}{2.5ex} Method → & MambaAD & ViTAD & Dinomaly & INP-Former & MVAD$^{\dagger}$ & \multirow{2}{*}{\textbf{Ours$^{\dagger}$}} \\ \cline{1-1}
\rule{0pt}{2.5ex} Category ↓ & NeurIPS 24 & CVIU 25 & CVPR 25 & CVPR 25 & TMM 26 & \\ \hline
\rule{0pt}{2.5ex}Audiojack & 84.2/21.6 & 80.6/21.3 & 86.8/48.1 & \underline{87.3}/\underline{49.4} & 82.9/30.8 & \textbf{87.1}/\textbf{49.5} \\ 
Bottle Cap & \underline{92.8}/30.6 & 82.2/16.7 & 89.9/\underline{32.4} & 87.3/28.6 & \textbf{93.9}/22.0 & 90.1/\textbf{32.6} \\ 
Button Battery & 79.8/46.7 & 77.0/\underline{49.1} & \textbf{86.6}/46.9 & 83.4/\textbf{54.3} & 78.9/49.6 & \underline{86.2}/44.8 \\ 
End Cap & 78.0/12.0 & 72.6/5.7 & 87.0/\underline{26.2} & 85.2/22.3 & 81.4/13.2 & \textbf{87.4}/\textbf{26.6} \\ 
Eraser & 87.5/30.2 & 86.0/24.8 & 90.3/39.6 & \underline{91.3}/\underline{41.4} & 89.3/27.8 & \textbf{91.5}/\textbf{45.6} \\ 
Fire Hood & 79.3/25.1 & 75.6/18.0 & 83.8/38.4 & \underline{84.6}/\underline{39.8} & 79.8/25.4 & \textbf{85.7}/\textbf{42.1} \\ 
Mint & 70.1/15.9 & 69.2/13.9 & \textbf{73.1}/\underline{22.0} & \textbf{73.3}/\textbf{24.4} & 69.9/16.0 & \underline{72.9}/\underline{22.2} \\ 
Mounts & 86.8/31.4 & 84.9/25.1 & \textbf{90.4}/39.9 & 89.1/\underline{38.8} & 88.5/29.4 & \textbf{90.4}/\textbf{41.6} \\ 
PCB & 89.1/46.3 & 83.3/25.7 & \underline{92.0}/\textbf{55.0} & \textbf{92.4}/53.5 & 89.8/42.3 & 91.8/\underline{53.8} \\ 
Phone Battery & 90.2/36.3 & 90.7/24.1 & \underline{92.9}/51.6 & \textbf{93.0}/\underline{56.0} & 91.3/25.5 & \underline{92.9}/\textbf{61.8} \\ 
Plastic Nut & 87.1/33.1 & 82.3/20.8 & 88.3/41.0 & \underline{88.9}/\underline{41.1} & 87.4/28.7 & \textbf{89.7}/\textbf{42.9} \\ 
Plastic Plug & 85.7/24.2 & 80.8/13.3 & \textbf{90.5}/\underline{31.7} & 89.6/29.8 & 86.7/24.3 & \underline{90.0}/\textbf{34.1} \\ 
Porcelain Doll & 88.0/31.3 & 86.4/21.6 & 85.1/27.9 & \textbf{89.1}/\textbf{35.3} & \textbf{89.1}/28.1 & \underline{88.2}/\underline{34.3} \\ 
Regulator & 69.7/20.6 & 61.3/7.6 & \textbf{85.2}/\underline{42.2} & 78.9/37.0 & 70.4/16.0 & \underline{83.6}/\textbf{44.0} \\ 
Rolled Strip Base & 98.0/37.4 & 98.3/27.1 & \textbf{99.2}/41.6 & 99.0/\underline{41.7} & 97.9/26.8 & \textbf{99.2}/\textbf{42.0} \\ 
SIM Card Set & 94.4/51.1 & 92.9/23.3 & \underline{95.8}/\underline{52.1} & \textbf{96.5}/\textbf{60.9} & 93.6/47.6 & 95.6/45.6 \\ 
Switch & 91.7/39.9 & 87.8/28.8 & \textbf{97.8}/\textbf{62.3} & 97.0/57.8 & 91.9/39.1 & \underline{97.6}/\underline{61.7} \\ 
Tape & 96.8/47.1 & 93.6/30.7 & 96.9/\underline{54.0} & \underline{96.9}/53.0 & \textbf{97.1}/40.3 & \textbf{97.5}/\textbf{55.4} \\ 
Terminal Block & 96.1/35.3 & 89.6/20.1 & \textbf{96.7}/\underline{48.0} & \underline{95.5}/47.8 & 94.8/32.4 & 95.4/\textbf{51.7} \\ 
Toothbrush & 85.1/27.8 & 81.8/25.3 & \textbf{90.4}/\textbf{38.3} & \underline{90.2}/33.5 & 85.5/27.7 & 89.8/\underline{38.4} \\ 
Toy & 83.0/16.4 & 76.5/9.2 & \textbf{85.6}/\textbf{22.5} & 84.3/\underline{21.4} & 82.4/14.7 & \underline{84.5}/\underline{21.6} \\ 
Toy Brick & 70.5/18.0 & 69.1/20.6 & 72.3/\underline{27.9} & \underline{72.4}/\textbf{33.8} & 70.1/17.2 & \textbf{74.5}/31.1 \\ 
Transistor1 & 94.4/39.4 & 91.7/32.0 & \textbf{97.4}/\underline{53.5} & 97.0/\underline{53.7} & 95.6/39.6 & \underline{97.3}/\textbf{54.3} \\ 
U Block & 89.7/37.8 & 84.3/30.3 & 89.9/\underline{41.8} & \underline{90.3}/\textbf{46.6} & 89.8/31.8 & \textbf{91.3}/\underline{49.1} \\ 
USB & 92.0/39.1 & 86.9/26.1 & 92.0/\textbf{45.0} & \textbf{93.0}/42.2 & \underline{92.4}/37.8 & 91.8/\underline{44.3} \\ 
USB Adaptor & 79.4/15.3 & 76.1/5.9 & 81.5/23.7 & \textbf{84.5}/\textbf{26.5} & 79.4/15.3 & \underline{82.8}/\textbf{26.5} \\ 
Vcpill & 88.3/50.2 & 82.2/44.4 & 92.0/\underline{66.4} & \textbf{92.6}/\textbf{68.7} & 83.4/43.1 & \underline{92.4}/\underline{67.4} \\ 
Wooden Beads & 82.5/32.6 & 82.8/31.0 & 87.3/45.8 & \underline{88.8}/\textbf{51.1} & 85.5/29.3 & \textbf{89.0}/\underline{49.2} \\ 
Woodstick & 80.4/40.1 & 75.6/38.4 & \underline{84.0}/\underline{50.9} & 83.0/\textbf{53.0} & 80.5/37.4 & \textbf{84.9}/\underline{52.4} \\ 
Zipper & 99.2/58.2 & 98.8/47.8 & \underline{99.1}/\underline{67.2} & \underline{99.1}/\textbf{68.5} & \textbf{99.2}/49.8 & \underline{99.1}/65.6 \\ \hline \rowcolor{cyan!8}
\rule{0pt}{2.5ex} \textbf{Average} & 86.3/33.0 & 82.7/24.3 & \underline{89.3}/42.8 & 89.1/\underline{43.7} & 86.6/30.3 & \textbf{89.7}/\textbf{44.4} \\ 
\specialrule{1.5pt}{0pt}{0pt}
\end{tabular}
\label{tab:a3}
\end{center}
\end{table}
